\documentclass[AMA,STIX1COL]{WileyNJD-v2}

\articletype{Article Type}%

\received{}
\revised{}
\accepted{}
        
\begin{document}

\title{ Feature Fusion Detector for Semantic Cognition of Remote Sensing}

\author[1]{Wei Zhou*, Yiying Li*}

\authormark{Wei Zhou \textsc{et al}}

\address{\orgdiv{College of Computer}, \orgname{National University of Defense Technology}, \orgaddress{\state{Changsha}, \country{China}}}

\corres{*Wei Zhou, Yiying Li. {liyiying10, zhouwei14}@nudt.edu.cn}

% \presentaddress{This is sample for present address text this is sample for present address text}

\abstract[Summary]{The value of remote sensing images is of vital importance in many areas and needs to be refined by some cognitive approaches. The remote sensing detection is an appropriate way to achieve the semantic cognition. However, such detection is a challenging issue for scale diversity, diversity of views, small objects, sophisticated light and shadow backgrounds. In this article, inspired by the state-of-the-art detection framework FPN \cite{Lin}, we propose a novel approach for constructing a feature fusion module that optimizes feature context utilization in detection, calling our system LFFN for Layer-weakening Feature Fusion Network. We explore the inherent relevance of different layers to the final decision, and the incentives of higher-level features to lower-level features. More importantly, we explore the characteristics of different backbone networks in the mining of basic features and the correlation utilization of convolutional channels, and call our upgraded version as advanced LFFN. Based on experiments on the remote sensing dataset from Google Earth, our LFFN has proved effective and practical for the semantic cognition of remote sensing, achieving 89\% mAP which is 4.1\% higher than that of FPN. Moreover, in terms of the generalization performance, LFFN achieves 79.9\% mAP on VOC 2007 and achieves 73.0\% mAP on VOC 2012 test, and advacned LFFN obtains the mAP values of 80.7\% and 74.4\% on VOC 2007 and 2012 respectively, outperforming the comparable state-of-the-art SSD and Faster R-CNN models.}


\maketitle

\section{Introduction}

In recent years, with the rapid development of sensor technology and aerospace remote sensing technology, the amount of information provided by remote sensing images is growing explosively, and this phenomenon also caters to the characteristics of big data. The value of remote sensing images needs to be refined via some cognitive approaches. For example, the semantic cognition of remote sensing can be achieved by the detection technologies, and such semantic cognition (e.g., the accurate semantic labels of objects in the accurate positions in remote sensing images) is of paramount importance to many intelligence applications, such as the poverty prediction \cite{Perez}, city planning \cite{cityP}, and archaeological discovery \cite{archaeology}. Because of the access to the semantic information, the intelligence systems related to remote sensing can provide meaningful cognitive services, and the quality of services mainly depends on the underlying detection technologies.

A main difference between the visible spectral remote sensing images and the ordinary images taken by cameras in our daily life is that the former is overlook maps showing a wide range of scenes from the high perspective. Consequently, as to the remote sensing images, the accurate detection of objects in multiple scales, especially the detection of small objects in a complex background (e.g., planes in large airports can be regarded as such small objects in remote sensing images) still remains a troublesome issue. For example, some small objects can not be located or some located small objects are hard to be classified.

\begin{figure}
	\centering
	\includegraphics[width=0.9\textwidth]{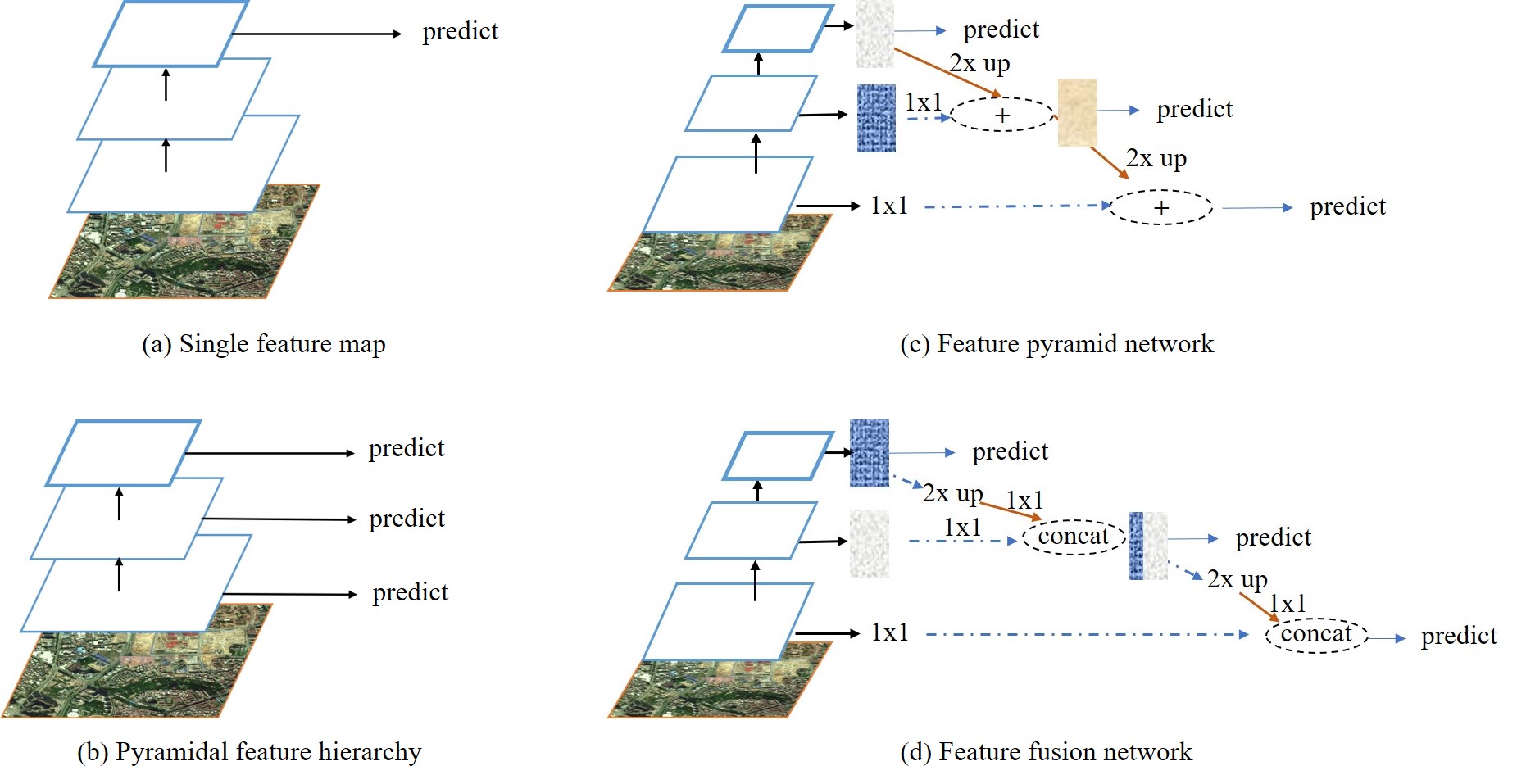}
	\caption{(a) Using only single scale features for the faster detection. (b) Some detection systems have reused the pyramidal feature hierarchy computed by a ConvNet. (c) Feature pyramid network is a top-down architecture with lateral connections \cite{Lin}. (d) Our proposed feature fusion network is as fast as (c), but more accurate. In this figure, feature maps are indicated by the rectangular structure with different colors representing the results of different levels.}
	\label{fig:1}       % Give a unique label
\end{figure}

Classically, the feature pyramids built upon image pyramids enable a model to detect objects across a large range of scales by scanning the model over pyramid levels \cite{image_pyramid}. This approach has been widely used in the era of hand-engineered features \cite{HOG} and indeed requires the dense scale sampling. Nowadays, with the development of deep learning, the convolutional neural network (CNN) is capable to represent higher-level semantics. However, the prediction via CNN based on a single topmost feature map (e.g., YOLO \cite{Redmon}) would lose some basic location information of objects (Figure 1(a)). Thus, a combination of CNN and the pyramids architectures comes as the Single Shot Detector (SSD) \cite{SSD}. SSD is the first to use a CNN pyramidal feature hierarchy which means to reuse the multiple feature maps from different layers of CNN (Figure 1(b)). Nevertheless, it misses the opportunity to reuse the feature hierarchy information, which will severely lose contextual information that is important to the detection of small objects. 

In a recent study, Feature Pyramid Network (FPN) \cite{Lin} provides a method to combine low-resolution, semantically strong features with the high-resolution, semantically weak features via a top-down pathway and lateral connections (Figure 1(c)). However, we have found that the indiscriminate injection of high-level features into low-level features in FPN will affect the detection accuracy of objects, and thus we hope to modify the connection degree of high-level features to the low-level features for the more reasonable utilization of feature pyramids relationship.

In this article, we propose a novel approach of constructing a feature fusion detector assigning differentiated stimulus degrees among different feature layers for the semantic cognition of remote sensing, named LFFN (Layer-weakening Feature Fusion Network). Our goal is to enable the network to learn to make better use of the feature interlayer relationship by itself instead of the pre-designed indiscriminate injections (i.e., direct adding operation) among layers. The feature pyramid method is important for the detection of objects in multi-scales because of the location and semantic information provided by different layers, and we aim to achieve the more effective feature pyramid structure via the novel feature fusion method. We change the non-differential inter-layer addition operations in the existing work to concat (e.g., Caffe concat layer \cite{Caffe}) operations that are decremented by layers (Figure 1(d)). In other words, we hope to integrate the features of different layers in a more reasonable way and then both the semantic and location information can be better utilized through the learning of the network. 

Our proposed system LFFN achieves the highest accuracy at present on the remote sensing dataset from Google Earth, surpassing all the existing state-of-the-art approaches. LFFN has proved effective and practical for remote sensing detection, achieving 89\% mAP (mean Average Precision) which is 4.1\% higher than that of FPN. Also, LFFN can be trained end-to-end with proper inference and training speed. In addition, we modify the basic backbone network (from the ResNet-50 \cite{ResNet} to the SE-ResNeXt-50 \cite{SENet}), construct the adaptive quantization module to increase the correlation utilization ratio of feature channels and integrate the new stochastic non-maximum suppression (NMS) to achieve the advanced LFFN with better performance. Moreover, our approach can also be applied generally on the standard object detection datasets, such as PASCAL VOC \cite{VOC}. LFFN achieves 79.9\% mAP on VOC 2007 and achieves 73.0\% mAP on VOC 2012 test, and advacned LFFN obtains the mAP values of 80.7\% and 74.4\% on VOC 2007 and 2012 respectively, outperforming a comparable state-of-the-art SSD and Faster R-CNN models. 

\section{Related Work}

Object detection in remote sensing images has been widely researched in recent years to achieve the valuable semantic cognition \cite{sensing}. Traditional methods mainly use local features to extract characteristics, such as Scale Invariant Feature Transform (SIFT) \cite{SIFT}, Histogram of Oriented Gradients (HOG) \cite{HOG}, and Saliency \cite{Saliency}. However, it is inconvenient to acquire the hand-engineered features and the feature extraction is insufficient. Recently, deep learning models have obtained increased attention in the object detection area, and those approaches can be roughly divided into two genres. The first is about two-phase approaches where a reasonable Region of Interest (ROI) is proposed in the first stage, and then the decision-making of the second stage is refined. Another genre is to eliminate the region proposal stage and directly train an end-to-end detector. These one-stage detectors are more accessible to train and more computationally efficient in applications. In the one-stage methods, multiple features utilization approaches can be used to improve the detection accuracy and our work is also based on the one-stage idea.

\textbf{Single feature map.} The first set of one-stage approaches, such as Faster R-CNN \cite{Faster_rcnn}, YOLO \cite{Redmon}, and R-FCN \cite{Dai}, use a bottom-up structure and only the topmost features for detection. YOLO converts the detection task into a regression problem \cite{Redmon}, significantly improving the speed of testing. At the same time, the global information is used in each prediction, so that the false positive rate is greatly reduced, but the detection accuracy is still not satisfactory enough. Moreover, such topmost feature map would lose some basic location information in low-level features, resulting in the limitations for detecting objects in different scales, especially for small objects in remote sensing images.

\textbf{Pyramidal feature hierarchy.} Another type of one-stage approaches utilizes different feature layers of backbone networks. Different feature layers have different scales, and each feature layer corresponds to one output result. However, there is no direct correlation between different feature layers and the output results are all independent. SSD \cite{SSD} combines regression ideas of YOLO with the anchor mechanism of Faster R-CNN \cite{Faster_rcnn} and returns results based on the multi-scale regional features of each position of the whole image. SSD has the good accuracy and real-time performance. However, it misses the opportunity to reuse the feature hierarchy information, which will severely lose contextual information that is important for the detection of small objects. Similar to SSD, MS-CNN \cite{MSCNN} predicts objects without combining the multiple layers of features hierarchy.

\textbf{Feature pyramid network.} To combine the information among different layers, FPN \cite{Lin} fuses low-level features with location information and high-level features with sufficient semantics, and it produces results from each pyramid layer. However, it is inefficient that high-level features are merged with low-level features by element-wise addition, and excessive high-level features can also affect the detection accuracy. A similar work is DSSD \cite{DSSD} which merges a large amount of contextual information based on the deconvolution layer. The deconvolution module integrates information from earlier feature maps and the deconvolution layers, and the integration strategy is the same as FPN. Therefore, how to better utilize the feature interlayer relationship for both the location information and semantics in the detection process is a core issue that we concern.

In fact, besides the design of detector structures, a reasonable underlying backbone network is also an important part for better extraction of features to achieve the satisfactory semantic cognition of images. As to the research of backbone networks, VGG \cite{VGG} and ResNet \cite{ResNet} have proven effective by increasing the depth of network. DenseNet \cite{DenseNet} and DPN \cite{DPN} can improve the learning and representation performance by adjusting the sequential connection mechanism among layers. Grouped convolutions can be used to increase the cardinality, as shown in ResNeXt \cite{ResNeXt}. The dependence of channels is modeled in SENet \cite{SENet} which re-calibrates the produced features based on the squeeze-and-excitation operations on the channels. Different backbone networks have different performances on the model size, computational complexity, and feature mining. Therefore, we also hope to leverage the suitable high-performance backbone network to improve the detection performance of the whole system.

\section{Layer-weakening Feature Fusion Network}
\label{LFFN:1}

Based on the structure of the Feature Pyramid Network (FPN) \cite{Lin}, our proposed Layer-weakening Feature Fusion Network (LFFN) integrates the novel feature fusion module among different layers to pass abundant context information (i.e., the location and semantic information) to the final prediction, learning to take full advantage of the feature interlayer relationship. The architecture overview of LFFN is shown in Figure 2. ResNet-50 is utilized as the backbone network (in grey) to extract the preliminary features. LFFN has the feature fusion module (in blue) that contributes to the significant improvement of the detection performance. The feature fusion module as elaborated in the following section learns to perform the context fusion of features from ResNet-50 among multiple layers. Each of the output feature layers of feature fusion module is used to predict the classification and bounding boxes regression of RPN. 

\begin{figure}
	\centering
	\includegraphics[width=0.85\textwidth]{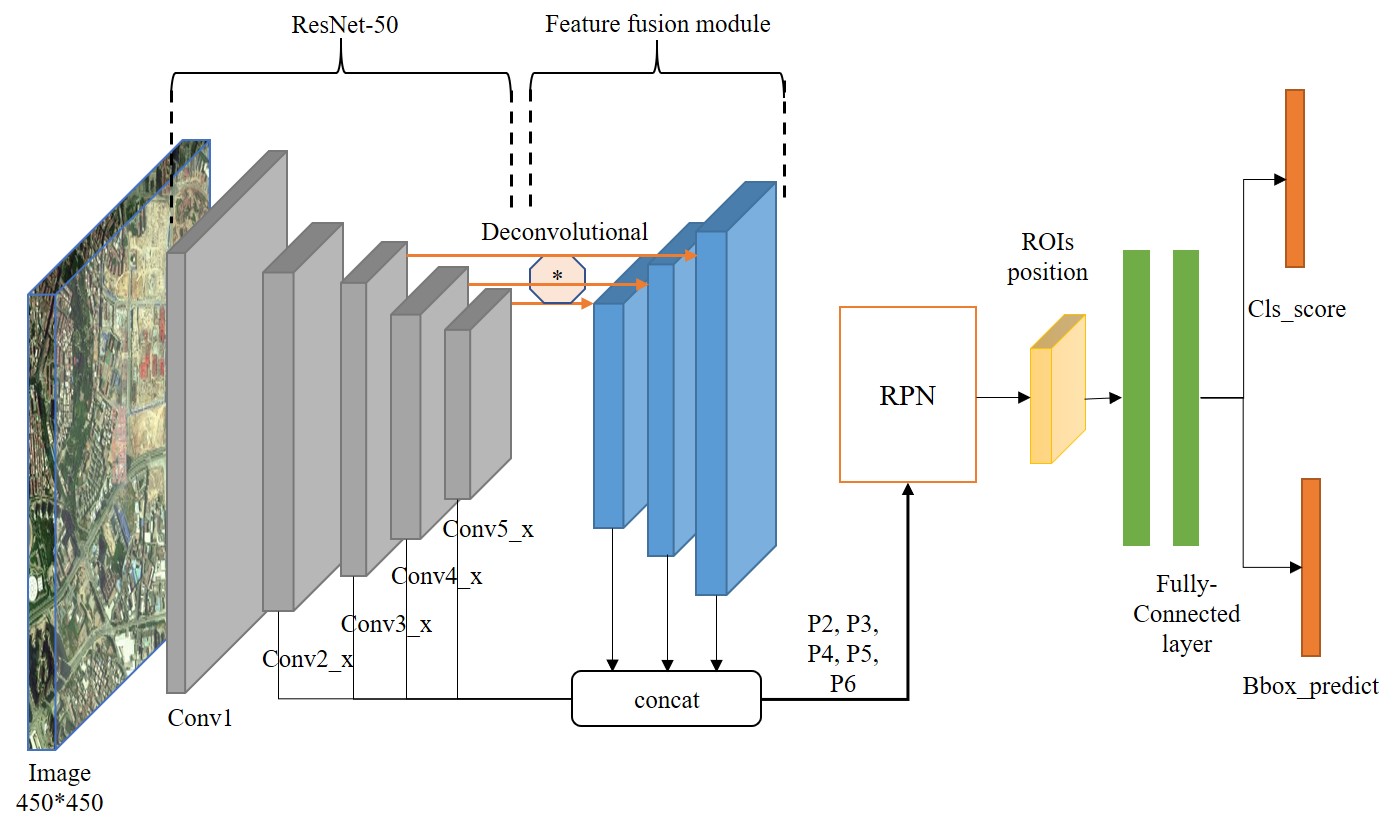}
	\caption{The architecture of LFFN.}
	\label{fig:2}       % Give a unique label
\end{figure}

\subsection{Feature Fusion Module of LFFN}
\label{LFFN:3}
 LFFN contains the basic backbone network for the extraction of features and the feature fusion module to learn to better explore the value of features among different layers. As shown in Figure 2, we find that the backbone network consists of five parts: conv1, conv2\_x, conv3\_x, conv4\_x, conv5\_x. For example, the output feature layer of conv4\_x can be regarded as a higher-level feature map than conv3\_x. These outputs will be the hierarchical inputs of our feature fusion module. The final set of feature maps to the RPN is \{P2, P3, P4, P5, P6\}, corresponding to \{conv2\_x, conv3\_x, conv4\_x, conv5\_x, conv5\_x max pooling\} that are respectively of the same resolution sizes. RPN is followed by two hidden 1,024-d fully-connected layers before the final classification and bounding box regression layers, and then the Non Maximum Suppression (NMS) \cite{NMS} is employed to post-process the predictions to get final semantic cognition (i.e., detection results).

The feature fusion module contains the layer-weakening top-down pathway and lateral connections. In our approach, such pathway leverages the deconvolution layer to upsample the higher level features. The feature fusion module is a computational unit which has a set of transformation functions, just like any given function  $Y=F_{fun}(X)$, $X\in \mathbb{R}^{H1\times W1\times C1}$, $Y\in \mathbb{R}^{H2\times W2\times C2}$. For simplicity, we concrete $F_{fun}$ into specific examples $F_{conv}$ and $F_{dev}$, which are the convolutional operator and deconvolutional operator. The kernel size of a simple convolutional layer is $(C_{out},C_{in},W,H) $, representing the output channels, input channels, kernel width, and kernel height, respectively. To make use of the information of different layers, we adopt an expansion-compression process as:

\begin{equation}
Z=F_{conv}(F_{dev}(X)))=\alpha (C_{conv}\ast \delta (D_{dec}^{T}\ast X))
\end{equation}
Here * denotes the convolution operation, $\alpha$ and $\delta$ refer to the ReLU functions. Expansion means that the convolutional layer is scaled from the plane space dimension using a deconvolution operation $D_{dec}$. The compression process $C_{conv}$ is to use the $1\times1$ convolution kernel to realize the dimension reduction.

In order to reduce the influence of high-level features, our upsampling process reduces the number of feature channels. By reducing the number of upsampling feature channels layer by layer, the overweighed influence of higher-level features on detection accuracy is then weakened. We achieve the cross-layer parameters sharing and preserve the intermediate features by connecting adjacent high- and low-level features in parallel, which can effectively reduce the feature redundancy by gradually weakening and reuse the existing features. That is where our work is significantly different from FPN, and it is also the reason for the better detection performance of objects in multiple scales.

\begin{figure}
	\centering
	\includegraphics[width=0.6\textwidth]{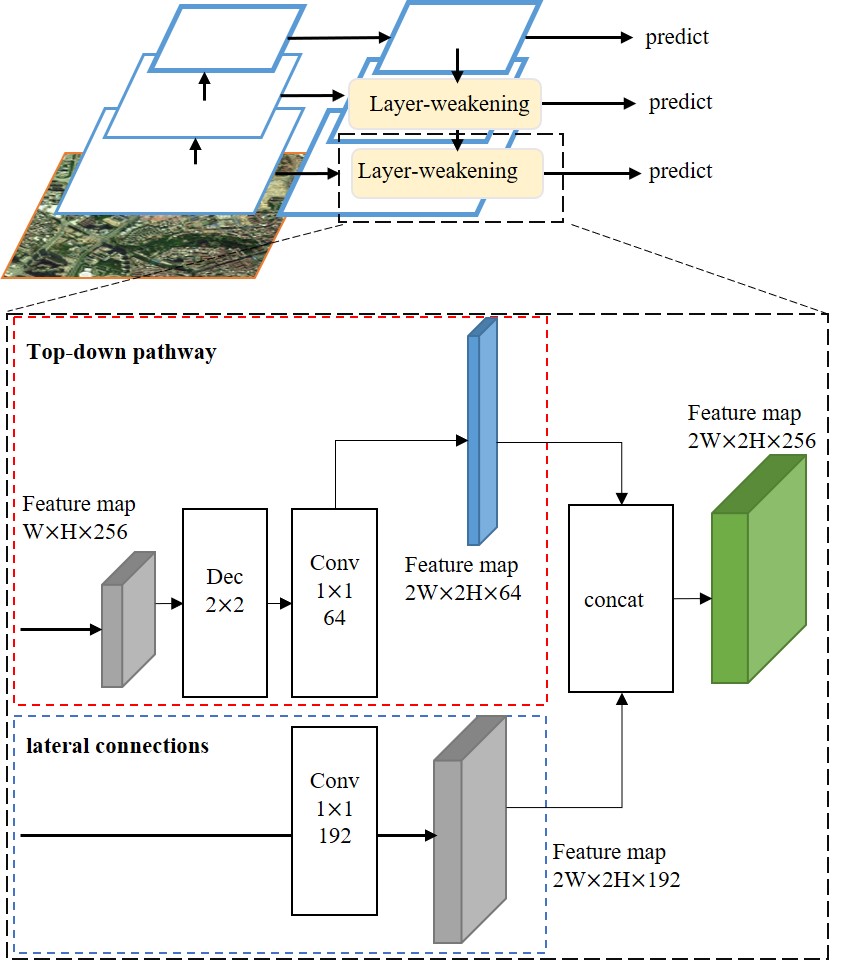}
	\caption{Feature fusion module that constructs the layer-weakening top-down pathway and the lateral connections.}
	\label{fig:3}       % Give a unique label
\end{figure}

Figure 3 shows the details of feature fusion module. We use a deconvolution layer to upsample the spatial feature maps by a factor of 2, and the upsampled feature maps are then processed by a 1*1 convolutional kernel to reduce the number of channels of high-level features. Meanwhile, to cater to the same channels number of output feature maps, we take a 1*1 convolution operation on the features derived from the bottom-up pathway. Each lateral connection merges feature maps produced from the bottom-up pathway with the upsampled feature maps by the concat operation. The concat process is as:

\begin{equation}
\left\{
\begin{array}{lr}
 Y\in \mathbb{R}^{H\times W\times C_{1}} \\
 Y^{'}\in \mathbb{R}^{H\times W\times C_{2}} \\
 Y_{out}=Y concat Y^{'} \in  \mathbb{R}^{H\times W\times (C_{1}+C_{2})}
 
\end{array}
\right.
\end{equation}
Then, these output feature maps transferred to RPN are {P2, P3, P4, P5, P6}. We set the number of channels of output feature maps as 256 in our work, and thus all other convolutional layers have 256-channel outputs.

\subsection{Transmitting Features to RPN}
The feature fusion module outputs more robust feature information by building context relations of multiple convolution hierarchies, and these feature information will be transmitted to RPN for bounding box proposal generation and object detection. To cover objects of different scales, the anchors of RPN have multiple predefined scales and aspect ratios \cite{Lin}. The feature maps that the feature fusion module passed to RPN are \{P2, P3, P4, P5, P6\}, which have different anchors due to different feature levels. The aspect ratios are \{0.5, 1, 2\}, and the scale is 8. With the settings, all anchor sizes are in \{32, 64, 128, 512, 1024\}. The label of the anchor part is positive under these two situations: one is when the anchor part has the largest Intersection over Union (IoU) with a ground-truth box; the other is when the anchor part has the IoU greater than 0.7 with any ground-truth. There is a case where the anchor tag is negative: all IoUs with the gound-truth boxes are less than 0.3. The final RPN outputs have a total of 128 positive and negative samples, and the corresponding ratio is 3:1. Then, we minimize the joint localization loss (e.g., Smooth L1) and the confidence loss (e.g., Softmax).

At the training stage, LFFN's backbone network ResNet-50 loads the pre-trained weight on the ImageNet1k classification set \cite{ImageNet}. Then, the entire neural network of LFFN will be fine-tuned on our training set.

\section{Advanced LFFN}
\label{ALFFN:1}
In this section, we mainly introduce our advanced LFFN which shows the improvements on the detection, considering not only the plane spatial but also the channel dimensions of ConvNet. Specifically, we first introduce the new backbone network SE-ResNeXt-50 in place of ResNet-50 for LFFN. Next we discuss how to add the adaptive quantization module to increase the correlation utilization ratio of channels outputted by the feature fusion module of LFFN, and how to integrate the new stochastic NMS to select the final frame for the advanced LFFN.
\subsection{Backbone Network of Advanced LFFN}
\label{ALFFN:2}
Convolutional neural networks have two spatial characteristics, namely the plane spatial and channel dimensions. The main focus of LFFN is to explore and utilize only the feature interlayer relationship in the plane spatial dimensions. Therefore, we hope to enhance the semantic cognition of LFFN by also taking the information on the channel dimensions into account. 

In line with this essential idea, SENet \cite{SENet} can learn the importance of each feature channel automatically, and then improve the useful features according to this importance and suppress features that are of little use to the current task. Thus, our first modification to LFFN is to use the SE-ResNeXt-50 \cite{SENet} as the backbone network in place of ResNet-50 used in the original LFFN. The SE-ResNeXt-50 effectively integrates existing networks and achieves complementary advantages, and its excellent performance has been proved in the three missions of image recognition, image detection and image segmentation\cite{SENet}. In our advanced LFFN, to solve the problems of the increase in model parameters and the heavy calculations caused by the new backbone network, we will adjust some super parameters, such as batch size or input image size, during subsequent experiments.

\subsection{Adaptive Quantization Module}

In addition to the modification of the backbone network, there are also differences in the system architecture of the advanced LFFN. Consistent with the idea of SENet on channels, we propose an adaptive quantization module. It is an adaptive optimized operator, which can construct strong context struture informations by fusing both plane spatial and channel-wise dimension (Figure 4). As introduced before, the basic LFFN can get the layer-weakening feature fusion outputs via the feature fusion module. The output of the feature fusion module is a convolution layer $Y\in \mathbb{R}^{H\times W\times C}$, $Y=[y_{1},y_{2},...,y_{C}]$ and $y_{i}$ is a 2D plane spatial structure ($ i$ indicates the channel number, and $i\in [1,C]$). In the system of advanced LFFN, in order to fully capture the channel-wise dependencies, we follow the basic feature fusion module with a serial operation called adaptive quantization module. The adaptive quantization module can automatically learn how to choose the quantization scale of the channel, and the relationship among channels is obtained through training. We parameterise the adaptive quantization module with one stochastic pooling layer \cite{stochasticNMS} , a fully connected (FC) layer and a Sigmoid function. As shown in Equation (3), a stochastic pooling process is generated by the probability $P(y_{i}(n,m))$ of each element of normalizing each channel spatial plane. The purpose of the stochastic pooling is to get better generalization performance. Then, an adaptive learnable relational network $W_{FC}$ is employed after the stochastic pooling process, $W_{FC} \in \mathbb{R}^{C\times C}$. $\beta$ refers to the Sigmoid function. At last, the final output $Y^{'}$ of the adaptive quantization module in the Equation (4) is computed by scaling the Sigmoid output with feature fusion module output. $G_{scale}$ are the probabilities of a set of channel-wise $Y$. Therefore, this module considers not only the plane spatial but also the channel dimensions of ConvNet on the system architecture for better semantic cognition of objects in the multiple scales.

\begin{figure}
	\centering
	\includegraphics[width=0.5\textwidth]{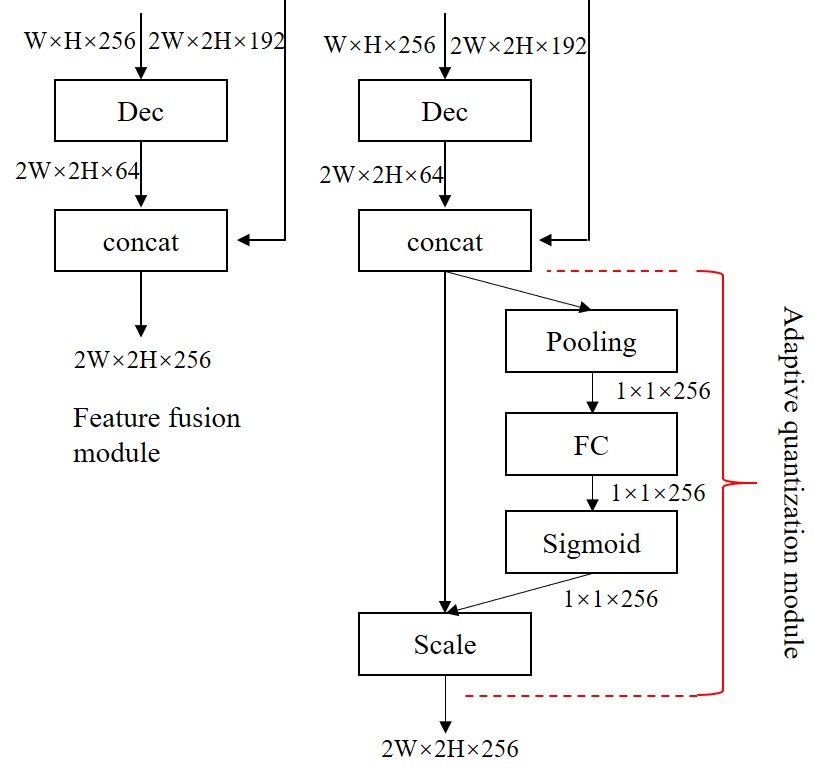}
	\caption{Comparison between the feature fusion module and the adaptive quantization module.}    
	% Give a unique label
\end{figure}

\begin{equation}
\left\{
\begin{array}{lr}

p_{i}=y_{i}(n,m) \quad  where \quad n,m\sim P(y_{i}(n,m)) \\
F=W_{FC}P \quad where \quad P=[p_{1},p_{2},..., p_{c}]  \\
G_{scale}=\beta F

\end{array}
\right.
\end{equation}

\begin{equation}
y^{'}_{c}=G_{scale}^{c} \cdot y_{c} 
\end{equation}

\subsection{Stochastic NMS}  
\label{ALFFN:3}

Non-maximum suppression (NMS) is an edge thinning method \cite{NMS}. It first generates a detection frame based on the object detection score, and then the detection frame M with the highest score is selected, while other detection frames that are significantly overlapped with the selected detection frame are suppressed. Here, we introduce a novel, simple and effective method called stochastic NMS for selecting the final frame. The stochastic NMS replaces the traditional NMS operations with a stochastic procedure, randomly picking the activation of each frame according to a possibility. The stochastic NMS is introduced to further improve the semantic cognition of objects. The traditional NMS uses a hard threshold to determine whether the adjacent detection frames are reserved. ${s}_{i}$ indicates the score of the ${i}$th detection frame ${b}_{i}$, and ${N}_{i}$ indicates the NMS detection threshold:

\begin{equation}
{{s}_{i}}=\left\{ 
\begin{array}{lr}
{{s}_{i}},{ iou(M,}{{{b}}_{i}}{)}\textless {{{N}}_{t}}  \\
0,{  iou(M,}{{{b}}_{i}}{)}\ge {{{N}}_{t}}  \\
\end{array} \right.
\end{equation}

As to our stochastic NMS method, the score reset function shows an effective improvement of the NMS method by probabilistically attenuating the detection scores of adjacent detection frames that overlap with the detection frame $M$. The smaller the area ratio between $iou(M,b_{i})$ and $b_{i}$, the less likely the frame is to be retained. More precisely, we first compute the probability $p$ for each frame $b_{i}$:
\begin{equation}
p_{i}=\frac{iou(M,b_{i})}{b_{i}}
\end{equation}
Then, we sample from the distribution based on $p_{i}$ to decide whether to retain the score of the frame $b_{i}$. The classification performance of some categories is improved according to our stochastic NMS.

\begin{equation}
{{s}_{i}}=\left\{ 
\begin{array}{lr}
{{s}_{i}},{ iou(M,}{{{b}}_{i}}{)}\textless {{{N}}_{t}}  \\
{{s}_{i}},{ where \quad i\sim P(p_{i}) \ and \ iou(M,}{{{b}}_{i}}{)}\ge {{{N}}_{t}}  \\
{0},{ else}

\end{array} 
\right.
\end{equation}

\section{Results and Discussion}
\label{result:1}
In this section, we analyze the performance of our feature fusion detector for semantic cognition in two parts. The first is to analyze the performance of our LFFN based on the remote sensing dataset from Google Earth. Then, to evaluate the generalization of our methods, we also take experiments on the generic datesets VOC 2007 and VOC 2012.

\subsection{Dataset and Evaluation Methodology}
\label{result:2}
We take the experiments on a NVIDIA 1080TI GPU and an IBM x3660 M4 server. We adjust the settings to find the appropriate image size, minibatch size to prevent the error of out of memory. Since the 1080TI graphics card has about 11G of video memory, and finally we make the program take 10G for training.

To evaluate the performance of LFFN, we employ the remote sensing dataset collected from Google Earth. The dataset contains four categories of objects (i.e., plane, bridge, storage, and harbor). There are 95637 training images and 2138 test images, with the resolution as 450*450. The whole network of LFFN is trained for 250,000 steps. The initial learning rate is set as 0.001, and the batch size is 1. In addition, we further conduct experiments on a common dataset PASCAL VOC dataset \cite{VOC} to evaluate its generalization performance in the semantic cognition of objects in multiple scales, especially for the performance of the advanced LFFN. For the VOC datasets, we use the 'trainval' set for training and 'test' set for testing. We train the models on the union set of VOC 2007 trainval and VOC 2012 trainval, and evaluate them on VOC 2007 and VOC 2012 test set. To analyze the performance, we adopt the precision and recall, which can judge how relevant a set of ranked results is for approaches. In addition, we record the Average Precision (AP) which is a metric of object detection performance.

\subsection{Computational Complexity Analysis}
\label{ALFFN:4}
To illustrate the cost of the backbone network of LFFN and advanced LFFN, we take the computation comparison between ResNet-50 and SE-ResNeXt-50. The ResNet-50 requires 3.86 GFLOPs in a single forward pass for a 224x224 pixel input image. However, the SE-ResNeXt-50 requires 3.98 GFLOPs, corresponding to a 0.3\% relative increase over the original ResNet-50 \cite{SENet}.

In practice, with a training mini-batch of 2 images, an iterative process involving forwards and backwards through ResNet-50 takes 2.006s, compared to 2.307s for SE-ResNeXt-50 (both are performed on a server with a NVIDIA 1080TI GPU). The test time for an image is 448ms and 577ms respectively.  SE-ResNeXt-50 introduces 2 million additional parameters beyond the 25 million parameters required by ResNet-50, corresponding to a 8\% increase.

\subsection{Results on Remote Sensing Dataset}
\label{result:3}

\begin{figure}
	\centering
	\includegraphics[width=0.75\textwidth]{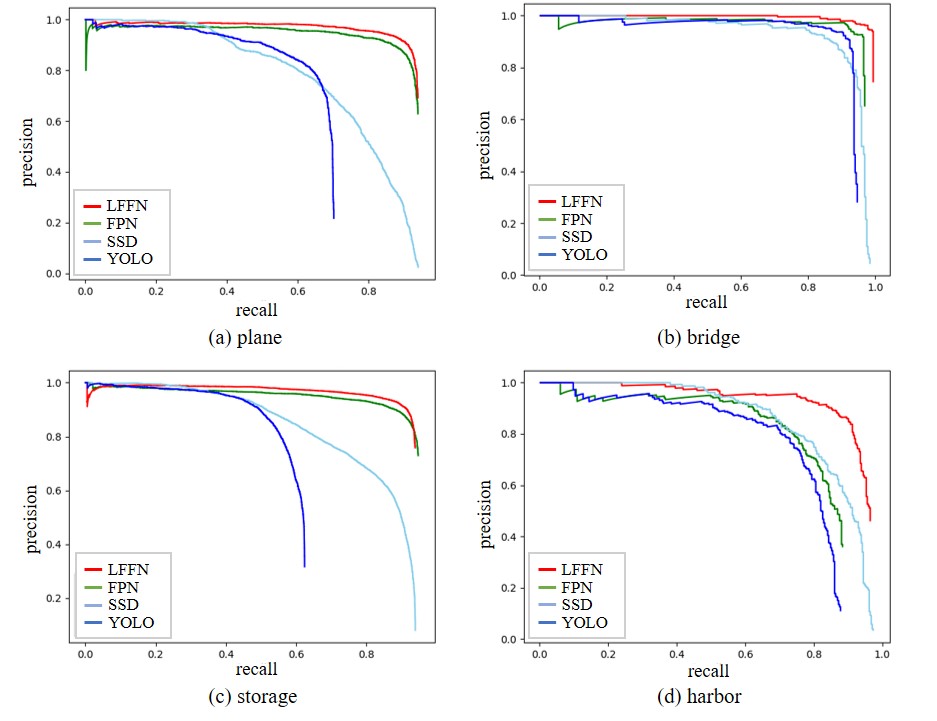}
	\caption{Precision-versus-recall curves of remote sensing detection.}
	\label{fig:3}       % Give a unique label
\end{figure}

For the comparison purpose, we use SSD \cite{SSD}, YOLO2 \cite{Redmon}, and FPN \cite{Lin} as horizontal comparisons to test the performance of semantic cognition (i.e., the detection accuracy). The object detection approaches return the bounding boxes with the classification scores. A detected bounding box is recognized as matched if the overlap with ground truth is larger than a certain value. The precision-versus-recall curves of these methods on the dataset from Google Earth are presented in Figure 5. It is clear that LFFN outperforms all others state-of-the-art methods by a considerable margin. The lower performance of the methods like the SSD and YOLO2 quantifies the difficulty of the detection task when the Google Earth dataset containing more small objects.

\begin{table}[]
	\centering
	\caption{Detection results on Google Earth dataset.}
	\label{my-label}
	%\resizebox{\textwidth}{!}{%
	\begin{tabular}{|l|l|l|l|l|l|l|}
			\hline
			{\textbf{Approach}} & {\textbf{Network}} & {\textbf{mAP}} & \multicolumn{4}{c|}{\textbf{AP}}   \\ \cline{4-7} 
			&                                   &                               & plane  & bridge & storage & harbor \\ \hline
			SSD\cite{SSD}                                & VGG16                             & 0.8045                        & 0.7316 & 0.8811 & 0.7853  & 0.8202 \\ \hline
			YOLO2\cite{Redmon}                              & Customized                        & 0.7105                        & 0.6358 & 0.8914 & 0.5844  & 0.7302 \\ \hline
			FPN\cite{Lin}                                & ResNet-50                         & 0.8475                        & 0.8704 & 0.8966 & 0.8723  & 0.7507 \\ \hline
			Our LFFN                           & ResNet-50                         & \textbf{0.8885} & \textbf{0.8901} & \textbf{0.9072} & \textbf{0.8894} & \textbf{0.8672} \\ \hline
	\end{tabular}%
	
\end{table}

The detection results, as shown in Table 1, validate that LFFN outperforms all others. LFFN achieves 88.9\% mAP which is 4.1\% higher than that of FPN and is much higher than that of SSD and YOLO2 (8.4\% and 17.8\% higher respectively). Notably LFFN is much better than other methods
which try to include context information such as FPN \cite{Lin}, even though LFFN simply takes a simple and effective context fusion process.

Figure 6 shows some detection examples of object categories with the confidence scores based on LFFN.  It can be seen that our approach detects most of the objects in multiple scales with high confidence. However, there are still some boundary occlusion and low-resolution objects. For example, some planes and harbors are missed in Figure 6(a). The main reasons are as follows: the objects and background are similar to a great extent, and in terms of small object, the loss of features after object occlusion has bad influence on object detection performance. We argue that occlusion follows a long-tail distribution, so the influence is limited.

\begin{figure}
	\centering
	\includegraphics[width=0.75\textwidth]{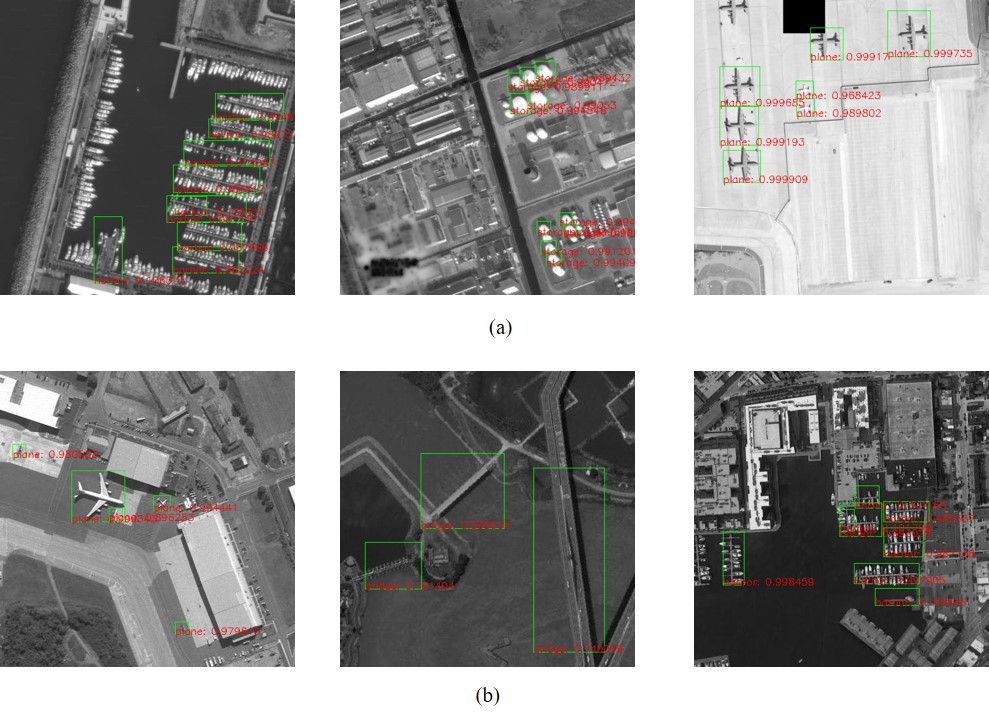}
	\caption{Detection examples on remote sensing dataset (e.g., plane, bridge, storage, harbor).}
	\label{fig:3}       % Give a unique label
\end{figure}

\subsection{Results on PASCAL VOC 2007}
\label{result:4}

% \usepackage{graphicx}

% \usepackage{graphicx}
\begin{table}[]
	\centering
	\caption{PASCAL VOC 2007 test detection results. Data: "07": VOC 2007 trainval, "07+12": union of VOC 2007 and VOC 2012 trainval. }
	\label{my-label}
	\resizebox{\textwidth}{!}{%
		$\begin{array}{l|l|l|llllllllllllllllllll}
		\hline
		\textbf{Approach}      & \textbf{data} & \textbf{mAP} & aero & bike & bird & boat & bottle & bus  & car  & cat  & chair & cow  & table & dog  & horse & mbike & person & plant & sheep & sofa & train & tv \\ \hline
		
		Fast \ R$-$CNN \cite{fastrcnn} & 07  & 66.9 &74.5& 78.3& 69.2& 53.2& 36.6& 77.3& 78.2& 82.0 &40.7 &72.7 &67.9& 79.6 &79.2 &73.0& 69.0& 30.1& 65.4& 70.2& 75.8& 65.8  \\ \cline{2-23} 
		
		Fast \ R$-$CNN \cite{fastrcnn} & 07+12 &70.0& 77.0 &78.1& 69.3& 59.4 &38.3 &81.6& 78.6& 86.7& 42.8& 78.8& 68.9& 84.7 &82.0& 76.6& 69.9 &31.8& 70.1 &74.8& 80.4& 70.4 \\ \cline{2-23} 
		
		Faster \ R$-$CNN \cite{Faster_rcnn} & 07 & 69.9& 70.0 &80.6 &70.1 &57.3 &49.9 &78.2& 80.4 &82.0& 52.2& 75.3 &67.2& 80.3 &79.8 &75.0& 76.3 &39.1 &68.3& 67.3& 81.1& 67.6 \\ \cline{2-23} 
		
		Faster \ R$-$CNN \cite{Faster_rcnn}                & 07+12              & 73.2         & 76.5 & 79.0 & 70.9 & 65.5 & 52.1   & 83.1 & 84.7 & 86.4 & 52.0  & 81.9 & 65.7  & 84.8 & 84.6  & 77.5  & 76.7   & 38.8  & 73.6 & 73.9 & 83.0 & 72.6  \\ \cline{2-23} 
		
		SSD \cite{SSD} & 07 & 68.0& 73.4& 77.5& 64.1 &59.0& 38.9& 75.2 &80.8& 78.5& 46.0& 67.8& 69.2& 76.6& 82.1& 77.0& 72.5& 41.2 &64.2& 69.1& 78.0& 68.5  \\ \cline{2-23}
		
		SSD \cite{SSD}                 & 07+12              &  74.3 &75.5 &80.2& 72.3& 66.3 &47.6 &83.0 &84.2& 86.1 &54.7& 78.3& 73.9& 84.5 &85.3& 82.6& 76.2 &48.6 &73.9& 76.0 &83.4 &74.0  \\ \hline 
		
		LFFN                   & 07        & 78.2         & 85.6 & 80.5 & 78.1 & 70.2 & 66.3   & \textbf{87.1} & 87.7 & 88.7 & 60.6  & 85.7 & \textbf{71.4}  & 87.5 & 86.7  & 79.9  & 79.4   & 48.2  & 79.4 & \textbf{79.0} & 85.6 & 75.9 \\ \cline{2-23}
		
		LFFN & 07+12 & 79.9 & \textbf{90.9} &81.8 & 80.9 & 71.8 & \textbf{72.0} & 81.8 & 80.8 &\textbf{90.9} & 54.5 & 80.8 & 71.4 & \textbf{90.9} &\textbf{88.9} &81.8 & \textbf{81.8} & 81.1 & 72.7 & 70.6 & \textbf{90.9} & \textbf{81.8}  \\ \hline 
		
		Advanced\ LFFN\ (with \ NMS)          & 07    & 78.9         & 86.2 & 86.3 & 78.8 & 73.1 & 66.0   & 83.7 & 88.0 & 89.3 & 64.1  & \textbf{88.3} & 68.6  & 87.5 & 87.3  & 80.1  & 78.9   & 54.0  & 78.9 & 76.0 & 83.1 & 79.5 \\ \cline{2-23}
		
		Advanced\ LFFN\ (with \ stochastic \ NMS) & 07    & 78.9         &86.6 & 86.5 & 78.9 & 73.1 & 65.8   & 84.0 & \textbf{88.1} & 89.3 & \textbf{64.4}  & 88.1 & 68.5  & 87.8 & 87.4  & 80.1  & 79.0   & 53.3  & 78.8 & 76.0 & 82.3 & 79.6  \\ \cline{2-23}
		
		Advanced\ LFFN\ (with \ NMS) & 07+12 & \textbf{80.7} & 90.7 & \textbf{86.5} & \textbf{81.5} &\textbf{73.3} & 71.4 & 84.2 & 81.0 &\textbf{90.9} & 54.3 & 81.4 & 71.1 & \textbf{90.9} &88.8 &81.8 & 81.6 & \textbf{81.2} & \textbf{80.8} & 70.2 & \textbf{90.9} & \textbf{81.8}  \\
			
		\hline
		\end{array}%
		$}
\end{table} 

To evaluate the generalization performance, we compare our methods against Fast R-CNN \cite{fastrcnn}, Faster R-CNN \cite{Faster_rcnn} and SSD on VOC 2007 test (4952 test images). SSD \cite{SSD} is the latest work with the new expansion data augmentation trick, and it has already been better than many other detectors.

Table 2 shows the test results on PASCAL VOC 2007. The 'data' in the table represents the training dataset. '07' indicates that the dataset is VOC 2007 trainval, and '07+12' indicates that the dataset is the union set of VOC 2007 trainval and VOC 2012 trainval. In general, in the case of larger dataset training, the approaches can obtain better performance. More importantly, by adding the feature fusion module, our LFFN is consistently better than that of Fast R-CNN, Faster R-CNN and SSD (9.9\%, 6.7\% and 5.6\% higher respectively) in the condition of '07+12' data. This proves the effectiveness of our proposed approach. 

Especially, the advanced LFFN reaches 80.7\% mAP, higher than that of LFFN, and the performance increases by nearly 1\%. Whether for LFFN or advanced LFFN, when using '07+12' training data, most APs of objects categories increase and the few decreased APs may lie in the over-fitting or inadequate training iterations for the more training data. We also make comparisons in the advanced LFFN when using the traditional NMS and our stochastic NMS. With a stochastic procedure, the system can randomly pick the activation of each frame based on a possibility, and then some categories showed changes in AP values. This is an interesting direction, and we will further explore it in the future. 

\subsection{Results on PASCAL VOC 2012}
\label{result:5}

For VOC 2012 task, we follow the setting of experiments on VOC 2007 and describe a few differences here. We use the whole VOC 2007 dataset and VOC 2012 trainval for training and VOC 2012 test for testing.

Table 3 shows the results of our LFFN and advanced LFFN, compared with other state-of-the-art methods. Our approaches can improve the performance of all the categories except bus, table and sofa in VOC 2012. We believe this is probably because of the larger diversity in VOC 2012. LFFN improves accuracy over Fast/Faster R-CNN, with 4.6\% and 2.6\%, respectively. Compared to YOLO and SSD, our approaches are more accurate, though both are faster than ours. LFFN and advanced LFFN based on more complex network, runs at 4 to 6 FPS, and these two are 50 PFS. Compared to LFFN, our advanced LFFN with SE-ResNeXt-50 and adaptive quantization module gives 1.4\% boost to 74.4\% mAP, acquiring better semantic cognition in images.

% \usepackage{graphicx}

\begin{table}[]
	\centering
	\caption{PASCAL VOC 2012 test detection results.}
	\label{my-label}
	\resizebox{\textwidth}{!}{%
		$\begin{array}{l|l|l|llllllllllllllllllll}
		\hline
		\textbf{Approach}      & \textbf{network} & \textbf{mAP} & aero & bike & bird & boat & bottle & bus  & car  & cat  & chair & cow  & table & dog  & horse & mbike & person & plant & sheep & sofa & train & tv \\ \hline
		
		YOLO\cite{Redmon}  & Customized & 57.9 & 77.0  & 67.2 & 57.7 & 38.3 & 22.7 & 68.3 &55.9 &81.4 &36.2 & 60.8 & 48.5 & 77.2 & 72.3 &71.3 & 63.5 & 28.9 & 52.2 & 54.8 & 73.9 &50.8 \\ \cline{2-23}
		
		Fast \ R$-$CNN \cite{fastrcnn} & VGG & 68.4 &82.3 &78.4 &70.8 & 52.3 & 38.7 &77.8 & 71.6 &89.3 &44.2 &73.0 &55.0 & 87.5 &80.5 &80.8 &72.0 & 35.1 & 68.3 & 65.7 &80.4 &64.2 \\ \cline{2-23}

		Faster \ R$-$CNN \cite{Faster_rcnn}  & ResNet$-$101  &70.4 &84.9 &79.8 &74.3 &53.9 & 49.8 & 77.5 &75.9 &88.5 & 45.6 & 77.1 & 55.3 & 86.9 &81.7 & 80.9 & 79.6 & 40.1 & 72.6 & 60.9 & 81.2 & 61.5  \\ \cline{2-23} 
		
		SSD \cite{SSD}     & VGG   & 72.4 & 85.6 &80.1 &70.5 & 57.6 &46.2 & \textbf{79.4} &76.1 & 89.2 & \textbf{53.0} & 77.0 & \textbf{60.8} & 87.0 & 83.1 & 82.3 & 79.4 & 45.9 & 75.9 &\textbf{69.5} & 81.9 & 67.5 
		\\ \hline

		LFFN                   & ResNet$-$50        & 73.0 & 85.4 & 81.9 & 74.4 & 58.5 & 58.2 & 76.8 & 77.5 & \textbf{90.7} & 48.8 & 77.9 & 56.0 & 87.9 & 83.0 & \textbf{82.8} & 82.4 & 52.6 & 74.5 & 62.6 & \textbf{82.1} & 65.4
		\\ \hline
		
		Advanced\ LFFN          & SE$-$ResNeXt$-$50    &  74.4 & \textbf{86.4} & \textbf{82.3} & \textbf{79.1} & \textbf{63.1} & \textbf{62.9} & 77.9 & \textbf{79.0} & 89.1 & 49.7 & \textbf{81.2} & 54.0 & \textbf{88.6} & \textbf{84.2} & 81.8 & \textbf{82.7} & \textbf{53.0} & \textbf{79.0} & 61.1 & 81.6 & \textbf{69.5}
		\\ \hline
		\end{array}%
		$}
\end{table}

\section{Conclusions}
\label{conlu:1}
In this article, we propose a progressive feature fusion detection approach LFFN for the semantic cognition in remote sensing images. It aims to address the troublesome detection problems like the scale diversity, diversity of views, small objects, and sophisticated light and shadow backgrounds. We explore the inherent relevance of different layers to the final decision, and the incentives of higher-level features to lower-level features. Moreover, we further explore the characteristics of different backbone networks in the mining of basic features, add the adaptive quantization module to increase the correlation utilization of convolutional channels, and integrate the new stochastic NMS to achieve the advanced LFFN system. Experiments show the high performance of our LFFN and advanced LFFN for acquiring semantic cognition in not only the remote sensing images but also the more general VOC dataset images. 

\section*{acknowledgements}
	
This work was supported in part by the National Natural Science Foundation of China under Grant 61202488 and 61751208.

\nocite{*}% Show all bib entries - both cited and uncited; comment this line to view only cited bib entries;
%\bibliography{wileyNJD-AMA}%

\end{document}